\documentclass[journal,twoside,web]{ieeecolor}

\usepackage{generic}
\usepackage{cite}
\usepackage{amsmath,amssymb,amsfonts}
\usepackage{graphicx}
\usepackage{url}
\usepackage{hyperref}
\hypersetup{hidelinks=true}
\usepackage{textcomp}

\graphicspath{{./figure/}{../figure/}}

\newcommand{\best}[1]{{\bfseries\boldmath #1}}
\newcommand{\missingfigure}[1]{%
  \fbox{\parbox[c][1.35in][c]{0.92\linewidth}{\centering Missing figure file: \texttt{#1}}}%
}

\begin{document}

\title{PROMISE-AD: Progression-aware Multi-horizon Survival Estimation for Alzheimer's Disease Progression and Dynamic Tracking}

\author{Qing Lyu, Jeremy Hudson, Mohammad Kawas, Yuming Jiang, Chenyu You, Christopher T Whitlow%
\thanks{Q. Lyu, J. Hudson, M. Kawas, and C. T. Whitlow are with the Department of Radiology and Biomedical Imaging, Yale School of Medicine, New Haven, CT 06510 USA.}%
\thanks{Y. Jiang is with the Department of Radiology Oncology, Wake Forest University School of Medicine, Winston-Salem, NC 27157 USA.}%
\thanks{C. You is with the Department of Applied Mathematics \& Statistics and
Department of Computer Science, Stony Brook University, Stony Brook, NY 11794 USA.}%
\thanks{Corresponding author: Qing Lyu (email: qing.lyu@yale.edu).}
\thanks{Code will be available after paper acceptance.}}

\markboth{\hskip25pc IEEE JOURNAL OF BIOMEDICAL AND HEALTH INFORMATICS}
{Lyu \MakeLowercase{\textit{et al.}}: PROMISE-AD for Alzheimer's Disease Progression}

\maketitle

\begin{abstract}
Individualized Alzheimer's disease (AD) progression prediction requires models that use irregular visits, account for censoring, avoid diagnostic leakage, and provide calibrated horizon risks. We propose PROgression-aware MultI-horizon Survival Estimation for Alzheimer's Disease (PROMISE-AD), a leakage-safe survival framework for predicting conversion from cognitively normal (CN) status to mild cognitive impairment (MCI) and from MCI to AD dementia using ADNI/TADPOLE tabular histories. PROMISE-AD converts pre-index visits into tokens containing standardized measurements, missingness masks, longitudinal changes, time-normalized slopes, visit timing, and non-diagnostic categorical attributes. A temporal Transformer fuses global, attention-pooled, and latest-visit representations to estimate a progression score and latent discrete-time mixture hazards. Training combines survival likelihood, horizon-specific focal risk loss, progression ranking, hazard smoothness, and mixture-balance regularization, followed by validation-set isotonic calibration for 1-, 2-, 3-, and 5-year risks. In held-out testing across three seeds, PROMISE-AD achieved an integrated Brier score (IBS) of 0.085 $\pm$ 0.012, C-index of 0.808 $\pm$ 0.015, and mean time-dependent AUC of 0.840 $\pm$ 0.081 for CN-to-MCI conversion, giving the lowest IBS among compared methods. For MCI-to-AD conversion, PROMISE-AD achieved the highest C-index (0.894 $\pm$ 0.018) and near-ceiling 5-year discrimination (AUROC 0.997 $\pm$ 0.003; AUPRC 0.999 $\pm$ 0.001), although some baselines had lower time-integrated Brier error. Ablation and interpretability analyses supported longitudinal change features, fused temporal representations, mixture hazards, cognitive and functional measures, APOE4 status, and recent conversion-proximal visits. These findings suggest that progression-aware survival modeling can provide interpretable multi-horizon AD conversion risk estimates.
\end{abstract}


\section{Introduction}
\IEEEPARstart{A}{lzheimer's} disease (AD) is a progressive neurodegenerative disorder whose clinical expression unfolds over years. Mild cognitive impairment (MCI) is commonly treated as an intermediate state between cognitively normal (CN) aging and dementia, but individuals with MCI follow heterogeneous trajectories: some convert rapidly to AD dementia, some remain stable, and others have uncertain follow-up due to censoring \cite{petersen1999mci}. Dynamic biomarker models further emphasize that AD-related pathology, neurodegeneration, cognition, and function evolve along partially overlapping temporal processes \cite{jack2010dynamic}. These properties make AD progression prediction a time-to-event problem rather than a static classification task.

Longitudinal cohorts such as the Alzheimer's Disease Neuroimaging Initiative (ADNI) have enabled systematic study of clinical, cognitive, genetic, imaging, and biochemical markers for early detection and progression tracking \cite{petersen2010adni}. The TADPOLE Challenge extended this effort by emphasizing prospective forecasting and identifying likely fast progressors for trial enrichment and care planning \cite{marinescu2019tadpole}. However, ADNI/TADPOLE data also expose central modeling challenges: visits are irregular, measurements are missing by design or availability, conversion events are sparse at early horizons, and subjects may be censored before the horizon of interest.

Many existing approaches handle only part of this structure. Binary classifiers can predict conversion within a fixed window, but they discard follow-up duration and require separate definitions for each horizon. Classical survival models, including the Cox proportional hazards model \cite{cox1972regression}, accommodate censoring but may be limited by proportional-hazards or linearity assumptions. More flexible survival approaches, such as random survival forests (RSF) \cite{ishwaran2008random}, gradient boosting \cite{chen2016xgboost}, and neural Cox or discrete-time survival models \cite{katzman2018deepsurv,lee2018deephit,gensheimer2019nnet}, relax some assumptions, yet many still rely on baseline or flattened summaries. Sequence models and attention mechanisms naturally represent repeated visits \cite{vaswani2017attention}; however, in standard form they are representation learners rather than time-to-event estimators: they do not account for right-censoring unless paired with a survival likelihood, do not produce calibrated probabilities at clinical horizons, and can learn shortcuts from diagnosis labels, stage variables, or post-index visits unless leakage controls are built into the data construction.

In this study, we propose PROMISE-AD, a PROgression-aware Multi-horIzon Survival Estimation framework for dynamic AD progression tracking from leakage-safe pre-index longitudinal histories. PROMISE-AD encodes each visit using numeric measurements, missingness indicators, longitudinal changes, slopes, timing information, and non-diagnostic categorical variables, while excluding diagnosis, stage variables, and post-index information. A temporal Transformer fuses global, attention-pooled, and latest-visit representations, then estimates a scalar progression score and latent discrete-time mixture hazards. Training combines survival likelihood, horizon-specific focal risk loss, progression ranking, hazard smoothness, and mixture-balance regularization, followed by validation-set isotonic calibration for 1-, 2-, 3-, and 5-year risks. The conceptual contribution is not any single module in isolation, but the task-specific coupling of leakage-safe AD conversion cohort construction, progression-aware visit tokenization, temporal representation learning, latent hazard-mixture modeling, and calibrated multi-horizon risk estimation. We evaluated PROMISE-AD on ADNI/TADPOLE-derived CN-to-MCI and MCI-to-AD conversion tasks using repeated subject-level splits and compared it with classical classifiers, survival models, sequence models, and modern tabular baselines across discrimination, calibration, Brier error, and risk-stratification metrics.

The main contributions of this work are fourfold. First, we design progression-aware visit tokenization for irregular clinical histories with incomplete records by jointly encoding observed measurements, missingness masks, changes from baseline, time-normalized slopes, visit timing, and non-diagnostic categorical attributes, allowing PROMISE-AD to explicitly handle missing values and longitudinally incomplete visits. Second, we introduce a hybrid temporal representation for subject-level risk estimation, in which a Transformer-based longitudinal encoder fuses a global classification token, an attention-pooled visit summary, and the latest pre-index visit embedding to balance long-term progression history with recent clinical state. Third, we develop latent mixture-based multi-objective survival learning for calibrated progression risk by combining discrete-time mixture hazards with survival likelihood, horizon-specific focal risk loss, pairwise progression ranking, hazard smoothness regularization, mixture-balance regularization, and isotonic calibration for 1-, 2-, 3-, and 5-year conversion probabilities. Fourth, we provide a comprehensive empirical and interpretability evaluation against classical classifiers, Cox-style models, random survival forests, recurrent and Transformer survival baselines, and modern tabular baselines, together with ablation and attribution analyses of longitudinal features, model components, and clinically relevant predictors.

\section{Related Study}

\subsection{Fixed-Horizon AD Progression Prediction and TADPOLE Benchmarks}
Early AD progression studies commonly framed prognosis as fixed-horizon classification, especially stable versus progressive MCI over 18--36 months. Structural MRI work showed that baseline atrophy, hippocampal and entorhinal measures, and whole-brain pattern classifiers could distinguish converters from non-converters \cite{cuingnet2011automatic,davatzikos2011prediction}. Multimodal SVM, multi-kernel, and related classifiers combined MRI, FDG-PET, CSF, cognition, APOE, and demographic variables, often improving discrimination over single-modality models \cite{hinrichs2011predictive,moradi2015mri}, while deep representation learning added latent MRI/PET features for AD/MCI classification and converter identification \cite{suk2014hierarchical}. Together, these studies established useful risk markers, but fixed-window designs collapse progression into heuristic converter labels.

The TADPOLE Challenge broadened ADNI benchmarking to prospective monthly forecasts of diagnosis, ADAS-Cog13, and ventricular volume \cite{marinescu2019tadpole}; its one-year follow-up compared 92 algorithms from 33 teams and emphasized prospective evaluation, heterogeneous longitudinal predictors, and uncertainty estimates \cite{marinescu2021tadpole}. Follow-on recurrent models used irregular visit histories for future clinical, cognitive, and ventricular forecasts \cite{nguyen2020minimalrnn}. These studies motivate PROMISE-AD's multi-horizon longitudinal setting, but they primarily evaluate scheduled outcome forecasting or fixed-window classification rather than calibrated time-to-conversion risk under right censoring, where follow-up duration, event timing, and censoring must be modeled explicitly.

\subsection{Survival Modeling for AD Conversion Risk}
Time-to-event methods explicitly represent follow-up duration and censoring. Cox proportional hazards regression remains foundational \cite{cox1972regression}, while random survival forests and gradient-boosted Cox models capture nonlinear effects and interactions \cite{ishwaran2008random,chen2016xgboost}. In AD-specific work, joint modeling of longitudinal markers and time-to-conversion has shown that cognitive, functional, and imaging trajectories improve prediction beyond baseline covariates, with ADAS-Cog and RAVLT often strong predictors \cite{li2017conversion}. Joint models couple repeated biomarker trajectories with event risk, but require parametric assumptions and can become complex with irregular, missing, multimodal predictors \cite{rizopoulos2012joint}. These approaches directly address censoring but often rely on baseline variables or engineered longitudinal summaries rather than end-to-end sequence representations.

\begin{figure*}[t]
    \centering
    \IfFileExists{figure/figure1.pdf}{%
        \includegraphics[width=5.75in]{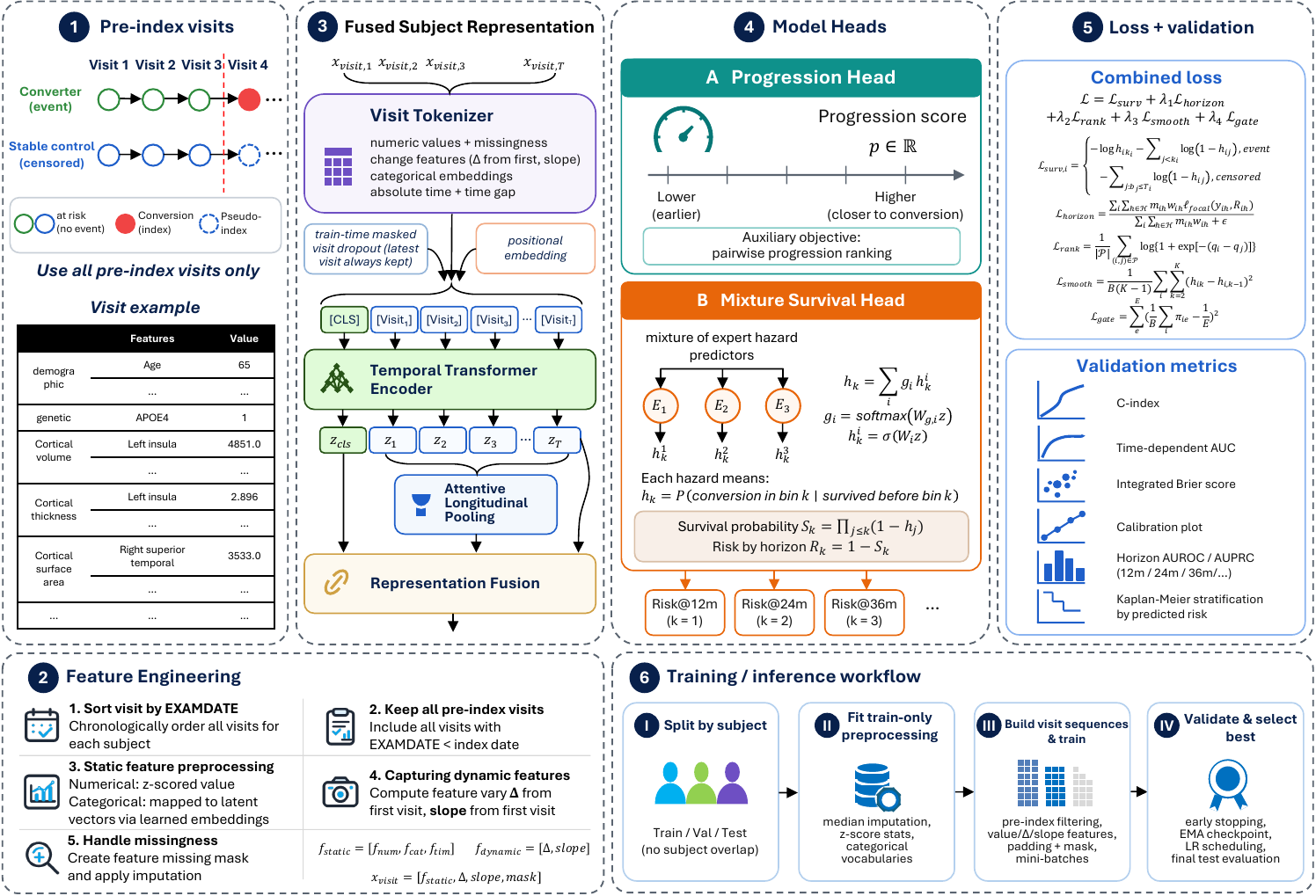}%
    }{%
        \missingfigure{figure/figure1.pdf}%
    }
    \caption{Overview of the PROMISE-AD framework, from leakage-safe pre-index visit construction and progression-aware tokenization to temporal encoding, latent mixture survival estimation, calibration, and multi-horizon evaluation.}
    \label{fig:figure1}
\end{figure*}

\subsection{Deep Longitudinal and Neural Survival Models}
Deep survival models learn nonlinear censored-risk functions: DeepSurv extends Cox modeling with neural representations \cite{katzman2018deepsurv}, CoxTime handles time-varying effects \cite{kvamme2019time}, and DeepHit-style discrete-time models estimate horizon-specific event distributions \cite{lee2018deephit,gensheimer2019nnet}, including longitudinal extensions such as Dynamic-DeepHit \cite{lee2020dynamicdeephit}. Modern tabular learners such as TabPFN-style summaries also provide strong comparison baselines for flattened predictors \cite{hollmann2023tabpfn}. However, neural survival models still require adaptation when inputs are irregular, incomplete pre-index visit sequences.

Recurrent neural networks are widely used for irregular clinical histories. LSTM, minimalRNN, and related AD models have predicted future diagnosis, cognition, ventricular volume, or progression stages from heterogeneous visits while addressing variable visit counts, uneven intervals, and missing data \cite{hochreiter1997long,wang2018rnn,nguyen2020minimalrnn,jung2021deeprecurrent}. Recent PPAD and TA-RNN architectures add recurrent, attention, time-embedding, or autoencoding components for irregular AD or EHR data \cite{alolaimat2023ppad,alolaimat2024tarnn}. These studies show the value of temporal neural networks, but their outputs are usually future states or biomarkers rather than survival-calibrated conversion probabilities.

Transformer encoders use attention to relate visits and features \cite{vaswani2017attention}, making them attractive for emphasizing earlier, recent, or dynamically changing AD visits. Yet generic sequence encoders do not address censoring, event-time likelihood, horizon labels, calibration, or leakage control by themselves. PROMISE-AD therefore links a temporal Transformer to discrete-time survival hazards, progression ranking, horizon focal loss, and post-hoc calibration for interpretable multi-horizon risk estimation.


\section{Methods}


\begin{table*}[t]
\centering
\tiny
\setlength{\tabcolsep}{1.5pt}
\caption{Comparison with baseline methods on held-out tests. For each task, C-index, IBS, 5-year AUROC, and 5-year AUPRC are shown; higher is better except IBS. Bold indicates the best value in each column.}
\label{tab:comparison}
\resizebox{\textwidth}{!}{%
\begin{tabular}{lcccc | cccc}
\hline
 & \multicolumn{4}{c}{CN-to-MCI} & \multicolumn{4}{c}{MCI-to-AD} \\
\cline{2-9}
Method & C-index & IBS $\downarrow$ & 5y AUROC & 5y AUPRC & C-index & IBS $\downarrow$ & 5y AUROC & 5y AUPRC \\
\hline
PROMISE-AD & $0.808\pm0.015$ & \best{$0.085\pm0.012$} & $0.864\pm0.023$ & $0.614\pm0.023$ & \best{$0.894\pm0.018$} & $0.095\pm0.007$ & \best{$0.997\pm0.003$} & \best{$0.999\pm0.001$} \\
\hline
CoxTime & $0.769\pm0.008$ & $0.188\pm0.013$ & $0.634\pm0.027$ & $0.360\pm0.038$ & $0.843\pm0.003$ & \best{$0.083\pm0.007$} & $0.981\pm0.012$ & $0.992\pm0.003$ \\
DeepHit & $0.722\pm0.012$ & $0.175\pm0.004$ & $0.720\pm0.049$ & $0.489\pm0.029$ & $0.838\pm0.011$ & $0.105\pm0.002$ & $0.963\pm0.014$ & $0.988\pm0.003$ \\
DeepSurv & $0.763\pm0.023$ & $0.188\pm0.021$ & $0.695\pm0.066$ & $0.409\pm0.061$ & $0.834\pm0.003$ & $0.092\pm0.004$ & $0.977\pm0.007$ & $0.991\pm0.002$ \\
Discrete MLP & $0.744\pm0.020$ & $0.174\pm0.011$ & $0.703\pm0.033$ & $0.469\pm0.053$ & $0.851\pm0.006$ & $0.093\pm0.008$ & $0.971\pm0.013$ & $0.989\pm0.003$ \\
RSF & $0.821\pm0.008$ & $0.140\pm0.001$ & $0.912\pm0.005$ & $0.690\pm0.016$ & $0.845\pm0.008$ & $0.131\pm0.000$ & $0.912\pm0.008$ & $0.978\pm0.003$ \\
XGBoost-Cox & \best{$0.843\pm0.000$} & $0.133\pm0.000$ & $0.911\pm0.000$ & \best{$0.711\pm0.000$} & $0.873\pm0.000$ & $0.110\pm0.000$ & $0.9716\pm0.000$ & $0.991\pm0.000$ \\
L2C-TabPFN & $0.768\pm0.000$ & $0.171\pm0.000$ & \best{$0.915\pm0.000$} & $0.705\pm0.000$ & $0.852\pm0.000$ & $0.094\pm0.000$ & $0.990\pm0.000$ & $0.994\pm0.000$ \\
LSTM & $0.811\pm0.016$ & $0.196\pm0.004$ & $0.868\pm0.008$ & $0.575\pm0.012$ & $0.866\pm0.005$ & $0.084\pm0.004$ & $0.989\pm0.001$ & $0.994\pm0.001$ \\
RNNAD & $0.808\pm0.018$ & $0.164\pm0.006$ & $0.773\pm0.025$ & $0.521\pm0.030$ & $0.861\pm0.002$ & $0.089\pm0.001$ & $0.977\pm0.006$ & $0.990\pm0.002$ \\
PPAD & $0.782\pm0.020$ & $0.178\pm0.005$ & $0.778\pm0.002$ & $0.502\pm0.012$ & $0.864\pm0.007$ & $0.085\pm0.003$ & $0.972\pm0.014$ & $0.990\pm0.003$ \\
TA-RNN & $0.812\pm0.030$ & $0.179\pm0.003$ & $0.783\pm0.078$ & $0.507\pm0.073$ & $0.860\pm0.001$ & $0.088\pm0.002$ & $0.986\pm0.005$ & $0.993\pm0.001$ \\
Seq Transformer & $0.775\pm0.024$ & $0.177\pm0.026$ & $0.734\pm0.072$ & $0.505\pm0.136$ & $0.848\pm0.008$ & $0.105\pm0.003$ & $0.927\pm0.051$ & $0.979\pm0.013$ \\
\hline
\end{tabular}
}
\end{table*}

\subsection{Problem Formulation and Cohort Construction}
We considered two time-to-conversion tasks: conversion from CN to MCI and conversion from MCI to AD dementia. For subject $i$, let
\begin{equation}
\mathcal{V}_i=\{(x_{i\ell}, c_{i\ell}, t_{i\ell})\}_{\ell=1}^{L_i}
\end{equation}
denote the retained pre-index visit history, where $x_{i\ell}$, $c_{i\ell}$, and $t_{i\ell}$ are numeric predictors, categorical predictors, and years since the first retained visit. Let $T_i$ denote time to conversion or control pseudo-index and $\delta_i$ indicate conversion. The objective is to estimate
\begin{equation}
F_i(h)=P(T_i\le h,\delta_i=1 \mid \mathcal{V}_i)
\end{equation}
for $h\in\{1,2,3,5\}$ years while ranking conversion-time risk under right censoring.

We used the ADNI TADPOLE D1/D2 longitudinal tabular dataset \cite{petersen2010adni,marinescu2019tadpole}, containing 12,741 visits from 1,737 participants and 1,907 columns before task-specific filtering. TADPOLE missing-value code $-4$ and blank strings were treated as missing.

Diagnostic stages from \texttt{DX\_bl}, \texttt{DX}, and \texttt{DXCHANGE} defined cohorts, event indicators, and index dates only. For converters, the index was the first valid post-baseline target diagnosis, and visits on or after it were excluded. Stable controls remained in the source state and received pseudo-index dates from post-baseline visits with at least two earlier visits, matched to converter follow-up time and pre-index visit count. Event or pseudo-index dates truncated histories and defined survival labels but were not model inputs. Rows with more than 10\% missingness, subjects with fewer than two retained pre-index visits, and inconsistent pre-conversion diagnoses were removed. Time was reset to the first retained visit, and subject-level 70\%/15\%/15\% splits were stratified by conversion label and pre-index follow-up.

\subsection{Leakage-safe Feature Engineering and Visit Tokenization}
Leakage control was applied at row, column, and preprocessing levels. Each token used only pre-index measurements; post-index values never contributed to measurements, missingness masks, changes, slopes, calibration, or model selection. Candidate predictors excluded information that could encode diagnosis or time-to-conversion, including \texttt{DX}, \texttt{DXCHANGE}, \texttt{baseline\_stage}, and \texttt{visit\_stage}. Remaining time features used only spacing among retained pre-index visits. Task-specific feature selection fixed columns before downstream splits, requiring at most 10\% missingness, sufficient variability, and no dominant mode above 99.5\%. The selection procedure prioritized clinically common variables and ensured representation from demographics, APOE/genetics, cognitive tests, cortical parcellation volumes, cortical surface area, cortical thickness, cortical thickness variability, white-matter or tissue volumes, and global MRI measures. This yielded 55 numeric columns for CN-to-MCI and 58 for MCI-to-AD; \texttt{PTGENDER} and \texttt{APOE4} were categorical.

All scaling, imputation, and categorical vocabularies were fitted on the training split only. Numeric predictors were median-imputed and standardized with training-set means and standard deviations. For numeric feature $p$ at visit $\ell$, PROMISE-AD encoded the standardized value, change from first retained visit, and time-normalized slope:
\begin{equation}
\begin{aligned}
z_{i\ell p} &= \frac{\tilde{x}_{i\ell p}-\mu_p}{\sigma_p},\\
\Delta z_{i\ell p} &= z_{i\ell p}-z_{i1p},\\
s_{i\ell p} &= \frac{\Delta z_{i\ell p}}{\max(t_{i\ell},10^{-3})}.
\end{aligned}
\end{equation}
The engineered numeric vector was concatenated with matching missingness masks, allowing the model to use imputed values without discarding missing-value information. Categorical variables used learned embeddings with special tokens for padding, unknown values, and missing values. Timing was represented by absolute time since the first retained visit and inter-visit gap. Let $r_{i\ell}$ denote the engineered numeric vector and $m_{i\ell}$ its missingness mask. The final visit token was
\begin{equation}
u_{i\ell}=\operatorname{LN}\{g_{\mathrm{num}}([r_{i\ell},m_{i\ell}])+
g_{\mathrm{cat}}(c_{i\ell})+
g_{\mathrm{time}}([t_{i\ell},\Delta t_{i\ell}])\},
\end{equation}
where $\operatorname{LN}$ denotes layer normalization and $g_{\mathrm{num}}$, $g_{\mathrm{cat}}$, and $g_{\mathrm{time}}$ are learnable projection networks.

\begin{figure*}[t]
    \centering
    \IfFileExists{figure/figure2.pdf}{%
        \includegraphics[width=5.75in]{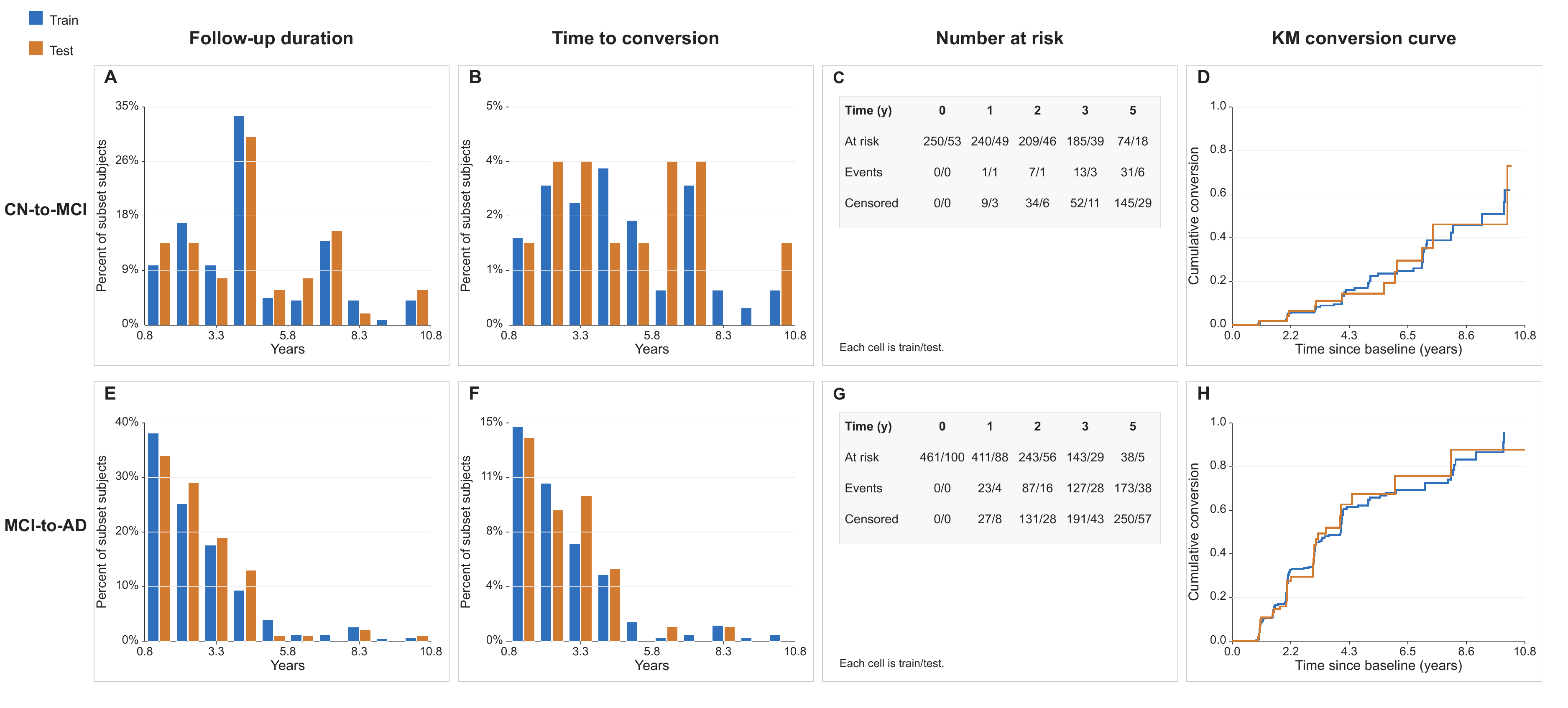}%
    }{%
        \missingfigure{figure/figure2.pdf}%
    }
    \caption{Cohort follow-up and conversion structure. Rows show CN-to-MCI and MCI-to-AD; columns show follow-up, event-time, at-risk/censoring, and Kaplan--Meier summaries.}
    \label{fig:figure2}
\end{figure*}

\subsection{PROMISE-AD Architecture}
PROMISE-AD applied masked visit dropout for irregular follow-up: each valid visit except the latest was independently dropped with probability $dp_v$. Retained tokens were prepended with a learned classification token, combined with positional embeddings, and encoded by a temporal Transformer \cite{vaswani2017attention}. The sequence was summarized by fusing the contextualized classification token, attention-pooled visit summary, and latest-visit embedding. A normalized multilayer perceptron produced subject representation $z_i$, progression score $q_i$, and discrete-time survival hazards.

Discrete-time hazards were modeled over bin endpoints $b=(0.5,1.0,1.5,2.0,2.5,3.0,4.0,5.0)$ years. To represent heterogeneous progression patterns, the survival head used a probability-space mixture of $E$ latent hazard experts. Expert gates, expert hazards, and final hazards were

\begin{equation}
\begin{aligned}
\pi_{ie} &= \operatorname{softmax}_e(W_gz_i),\\
h_{iek} &= \sigma(\eta_{iek}),\\
h_{ik} &= \sum_{e=1}^{E}\pi_{ie}h_{iek},
\end{aligned}
\end{equation}
where $\pi_{ie}$ is a soft membership weight and $h_{iek}$ is the expert-specific conditional hazard. Experts model latent rapid-conversion, delayed-progression, or lower-risk hazard patterns rather than predefined biological subtypes, yielding personalized mixture hazards. Survival and cumulative risk were

\begin{equation}
S_i(b_k)=\prod_{j=1}^{k}(1-h_{ij}),\quad
F_i(b_k)=1-S_i(b_k).
\end{equation}

For a requested horizon $h$, PROMISE-AD reported $F_i(b_k)$ at the first bin endpoint $b_k\ge h$.

\subsection{Training, Calibration, and Implementation}
The training objective combined five terms. Let $k_i$ denote the bin containing $T_i$. The discrete-time survival loss accounts for observed events before the final bin and for censored or later-event subjects:

\begin{equation}
\mathcal{L}_{\mathrm{surv},i}=
\begin{cases}
-\log h_{ik_i}-\sum_{j<k_i}\log(1-h_{ij}), & \text{event}\\
-\sum_{j:b_j\le T_i}\log(1-h_{ij}), & \text{censored}
\end{cases}
\end{equation}
where $j$ indexes discrete hazard intervals, $k_i$ is the interval containing the observed event time, and $b_j$ denotes the endpoint of interval $j$; thus the summations accumulate survival probabilities before conversion or censoring.
Observed events were upweighted by clipped inverse-frequency weights. We added focal risk loss at $\mathcal{H}=\{1,2,3,5\}$ years. With $R_{ih}=F_i(h)$, horizon label $y_{ih}$, evaluability mask $m_{ih}$, and class weight $w_{ih}$,

\begin{equation}
\begin{aligned}
\ell_{\mathrm{focal}}(y,R)
&=-y(1-R)^\gamma\log R-(1-y)R^\gamma\log(1-R),\\
\mathcal{L}_{\mathrm{horizon}}
&=\frac{\sum_i\sum_{h\in\mathcal{H}}m_{ih}w_{ih}
\ell_{\mathrm{focal}}(y_{ih},R_{ih})}
{\sum_i\sum_{h\in\mathcal{H}}m_{ih}w_{ih}+\epsilon}.
\end{aligned}
\end{equation}

The pairwise ranking term encourages higher progression scores for subjects who convert earlier. For comparable pairs $\mathcal{P}=\{(i,j):\delta_i=1,\ T_i<T_j\}$,

\begin{equation}
\mathcal{L}_{\mathrm{rank}}=
\frac{1}{|\mathcal{P}|}\sum_{(i,j)\in\mathcal{P}}
\log\{1+\exp[-(q_i-q_j)]\}.
\end{equation}

Hazard smoothness discourages abrupt adjacent-bin changes, and gate balancing discourages collapse to a single subtype expert:

\begin{equation}
\begin{aligned}
\mathcal{L}_{\mathrm{smooth}}
&=\frac{1}{B(K-1)}\sum_i\sum_{k=2}^{K}(h_{ik}-h_{i,k-1})^2,\\
\mathcal{L}_{\mathrm{gate}}
&=\sum_{e=1}^{E}\left(\frac{1}{B}\sum_i\pi_{ie}-\frac{1}{E}\right)^2 .
\end{aligned}
\end{equation}
Here, $B$ denotes the mini-batch size and $K$ is the number of discrete hazard intervals.

The full objective was

\begin{equation}
\begin{aligned}
\mathcal{L}={}&
\mathcal{L}_{\mathrm{surv}}
+\lambda_h\mathcal{L}_{\mathrm{horizon}}
+\lambda_p\mathcal{L}_{\mathrm{rank}}\\
&+\lambda_s\mathcal{L}_{\mathrm{smooth}}
+\lambda_g\mathcal{L}_{\mathrm{gate}}.
\end{aligned}
\end{equation}

PROMISE-AD was implemented in PyTorch and trained separately for CN-to-MCI and MCI-to-AD across seeds 0, 1, and 2. The model used $d_{\mathrm{model}}=128$, two Transformer layers, four attention heads, feed-forward dimension 256, dropout 0.25, four mixture experts, categorical embedding dimension 16, maximum sequence length 32, and visit dropout 0.15. Training used AdamW \cite{loshchilov2019decoupled} with learning rate $2\times10^{-4}$, weight decay $3\times10^{-4}$, batch size 32, gradient clipping at norm 1.0, and at most 180 epochs. Loss weights were empirically set as $\lambda_h=0.40$, $\lambda_p=0.10$, $\lambda_s=0.01$, and $\lambda_g=0.01$, with focal exponent $\gamma=1.5$. Checkpoints and early stopping were based on the total validation loss using exponential moving average weights with decay 0.995 and patience 30. Horizon risks were calibrated on the validation set using isotonic regression \cite{zadrozny2002transforming}; validation-set Youden thresholds were then applied for reported horizon classification metrics. All PROMISE-AD experiments were run on a single NVIDIA RTX PRO 6000 Blackwell GPU with 96 GB memory.


\begin{figure*}[t]
    \centering
    \IfFileExists{figure/figure3.pdf}{%
        \includegraphics[width=5.75in]{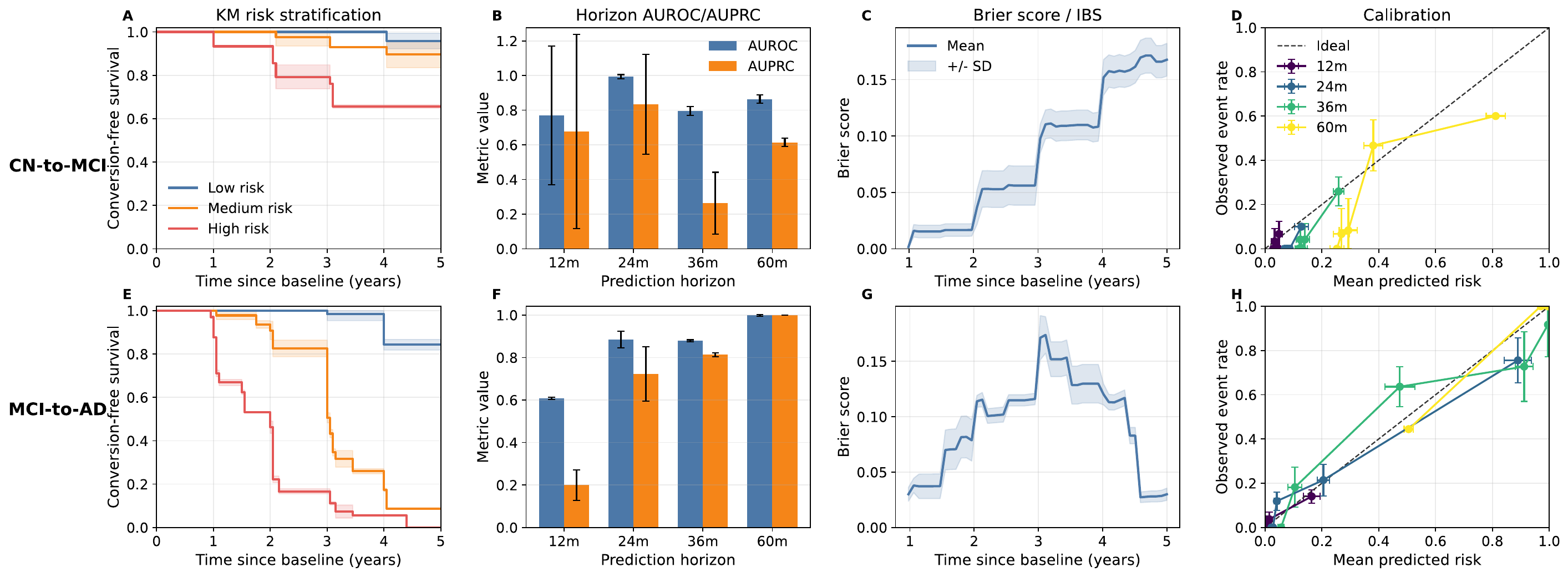}%
    }{%
        \missingfigure{figure/figure3.pdf}%
    }
    \caption{PROMISE-AD test performance. Rows show CN-to-MCI and MCI-to-AD; columns show risk-group survival (A, E), horizon AUROC/AUPRC (B, F), Brier score (C, G), and calibration (D, H).}
    \label{fig:figure3}
\end{figure*}

\section{Results}

\subsection{Dataset Analysis}
The final CN-to-MCI cohort included 358 subjects and 1,589 pre-index visits, split into 250/55/53 train/validation/test subjects with similar conversion rates (20.0\%, 21.8\%, and 22.6\%). Median follow-up was approximately four years in all splits (4.03/4.04/4.02 years for training/validation/testing), with a median of four visits per subject. CN-to-MCI events were sparse and late: only one test conversion occurred before 1 year and six before 5 years. This event pattern is a fundamental dataset constraint; fixed-horizon CN-to-MCI AUROC, AUPRC, and calibration estimates, including 5-year summaries, are therefore reported for completeness and should be interpreted as exploratory rather than stable early-progression evidence. The MCI-to-AD cohort included 661 subjects and 2,290 visits, split into 461/100/100 subjects with preserved conversion rates (40.6\%, 40.0\%, and 40.0\%). Median follow-up was approximately two years (2.01/2.01/2.01 years for training/validation/testing), with three visits per subject, and conversion occurred earlier and more frequently: 4, 16, 28, and 38 test conversions were observed before 1, 2, 3, and 5 years, respectively. Figure~\ref{fig:figure2} summarizes the resulting follow-up, event, censoring, and Kaplan--Meier structures.

\begin{figure*}[t]
    \centering
    \IfFileExists{figure/figure4.pdf}{%
        \includegraphics[width=5.25in]{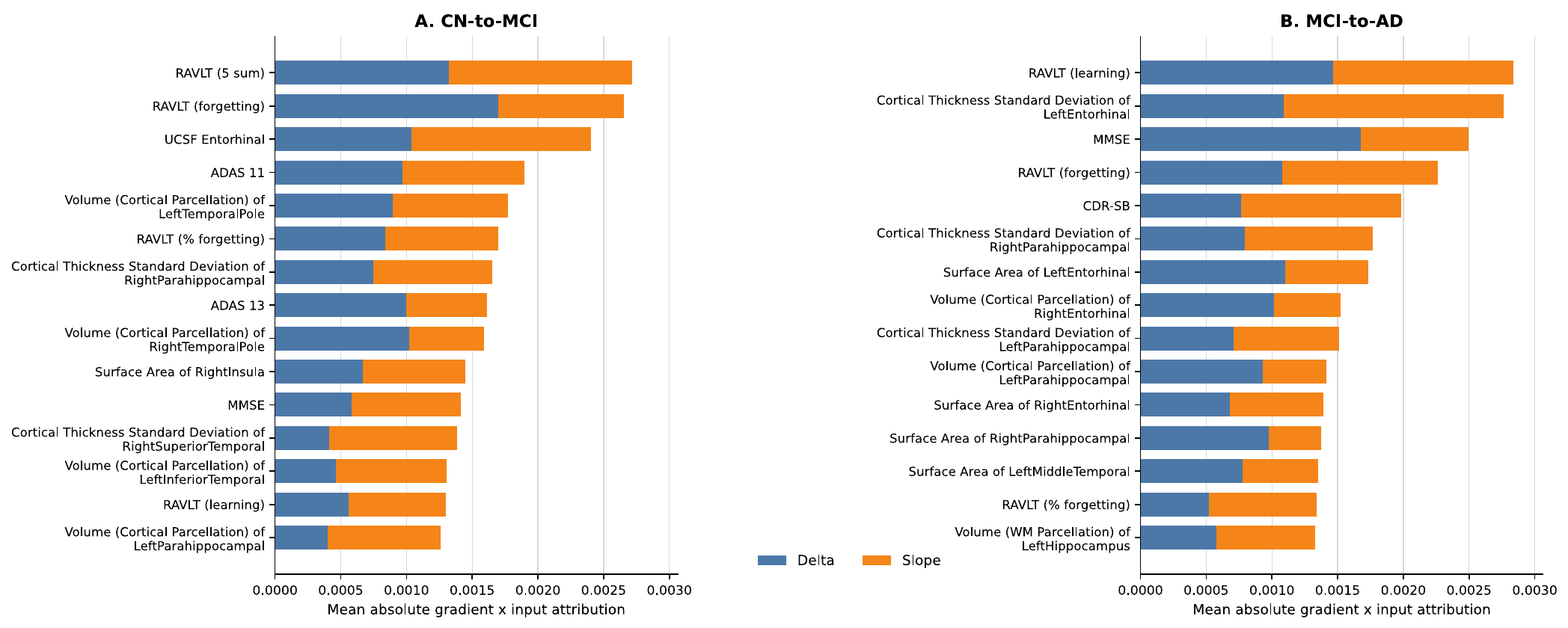}%
    }{%
        \missingfigure{figure/figure4.pdf}%
    }
    \caption{Top dynamic feature attributions for CN-to-MCI and MCI-to-AD.}
    \label{fig:figure4}
\end{figure*}

\subsection{PROMISE-AD Performance}
PROMISE-AD was evaluated separately for CN-to-MCI and MCI-to-AD on the held-out test subsets, using the best validation-loss checkpoint from each seed. Unless otherwise stated, all values below are reported as mean $\pm$ standard deviation across the three seeds. We summarized survival ranking with Harrell's concordance index \cite{harrell1982evaluating}, time-dependent discrimination with cumulative/dynamic AUC \cite{heagerty2000time}, probabilistic error with Brier score and integrated Brier score (IBS) \cite{brier1950verification,graf1999assessment}, and risk-group separation with Kaplan--Meier curves \cite{kaplan1958nonparametric}.

For CN-to-MCI, PROMISE-AD achieved C-index $0.808\pm0.015$ and IBS $0.085\pm0.012$ (Table~\ref{tab:comparison}). Because the test set contained only one converter before 12/24 months and six before 60 months, fixed-horizon AUROC/AUPRC estimates were event-limited; the 5-year AUROC $0.864\pm0.023$ and AUPRC $0.614\pm0.023$ among 24 evaluable subjects are reported only as exploratory summaries. In this task, time-integrated Brier error and concordance provide more stable evidence than early-horizon classification metrics. For MCI-to-AD, performance was stronger and more stable, with C-index $0.894\pm0.018$, IBS $0.095\pm0.007$, and near-ceiling 5-year discrimination (AUROC $0.997\pm0.003$, AUPRC $0.999\pm0.001$), reflecting the larger number of observed converters.

Figure~\ref{fig:figure3} shows that predicted risk groups were ordered in both tasks, with clearer separation for MCI-to-AD. CN-to-MCI calibration was limited by sparse events and some 60-month overestimation in the highest-risk bins, whereas MCI-to-AD showed strong risk stratification and better 5-year calibration. Overall, PROMISE-AD showed suggestive long-horizon CN-to-MCI risk separation, but the sparse event count prevents strong fixed-horizon conclusions; results were more stable for MCI-to-AD, consistent with the event-density differences in Figure~\ref{fig:figure2}.

\subsection{Comparison With Existing Methods}
We compared PROMISE-AD with survival, tabular, and sequence baselines, including CoxTime \cite{kvamme2019time}, DeepHit \cite{lee2018deephit}, DeepSurv \cite{katzman2018deepsurv}, discrete-time survival MLP \cite{gensheimer2019nnet}, RSF \cite{ishwaran2008random}, XGBoost-Cox \cite{chen2016xgboost}, L2C-TabPFN \cite{hollmann2023tabpfn}, recurrent models \cite{hochreiter1997long,wang2018rnn,nguyen2020minimalrnn}, PPAD/TA-RNN \cite{alolaimat2023ppad,alolaimat2024tarnn}, and sequence Transformers \cite{vaswani2017attention}.

For CN-to-MCI, XGBoost-Cox achieved the highest C-index and 5-year AUPRC, while L2C-TabPFN achieved the highest 5-year AUROC. Because the CN-to-MCI 5-year comparison was based on six converters, these AUROC/AUPRC rankings should be treated as exploratory. PROMISE-AD's clearest advantage was probabilistic accuracy: its IBS was the lowest overall and reduced integrated Brier error relative to the best comparison-method IBS by approximately 36\%. For MCI-to-AD, PROMISE-AD achieved the highest C-index and strongest 5-year AUROC/AUPRC, whereas CoxTime achieved the lowest IBS. Thus, within the event-limited CN-to-MCI setting, PROMISE-AD's strongest evidence was lower time-integrated Brier error, while for MCI-to-AD it provided the strongest ranking and late-horizon discrimination.

\subsection{Ablation Study}
We ablated six design choices on MCI-to-AD: dynamic features, visit dropout, representation fusion, mixture hazards, progression ranking, and horizon focal loss (Table~\ref{tab:ablation-study}).

The full model achieved the best C-index, IBS, and 5-year discrimination in the ablation study (Table~\ref{tab:ablation-study}). Removing dynamic change/slope features produced the largest C-index drop and reduced 5-year AUROC/AUPRC, although it slightly increased mean time-dependent AUC. Removing representation fusion or mixture hazards degraded IBS and 5-year discrimination, supporting the value of combining global, pooled, latest-visit, and heterogeneous hazard information. The IBS increase after removing mixture hazards ($0.095$ to $0.107$) suggests that soft latent hazard patterns improved probability estimation, but this result should be interpreted as evidence for modeling risk heterogeneity rather than proof of clinically distinct subtypes. Visit dropout, progression ranking, and horizon loss had smaller but consistent effects.

\subsection{Model Interpretability}
We assessed interpretability using gradient-times-input attributions for dynamic features and visit-level attention weights from the representation-fusion module.

The mixture hazards provide an additional, model-level interpretation of heterogeneity. Each expert represents a latent hazard-time profile, and each subject receives soft gate weights over these profiles rather than a hard subtype label. Clinically, this structure is most naturally interpreted as separating rapid, delayed, and lower-risk survival patterns, which is consistent with known heterogeneity in AD progression. However, because no external subtype labels or independent biomarker-defined progression subtypes were used, the learned experts should be viewed as risk-pattern components rather than validated clinical subtypes.

Dynamic attributions concentrated on clinically plausible memory, cognitive, functional, and neuroanatomical markers (Figure~\ref{fig:figure4}), including RAVLT, ADAS, MMSE, CDRSB, entorhinal, parahippocampal, temporal, and hippocampal measures. These patterns indicate that PROMISE-AD used within-subject changes and slopes rather than only static baseline severity.

\begin{figure}[t]
    \centering
    \IfFileExists{figure/figure5.pdf}{%
        \includegraphics[width=0.85\columnwidth]{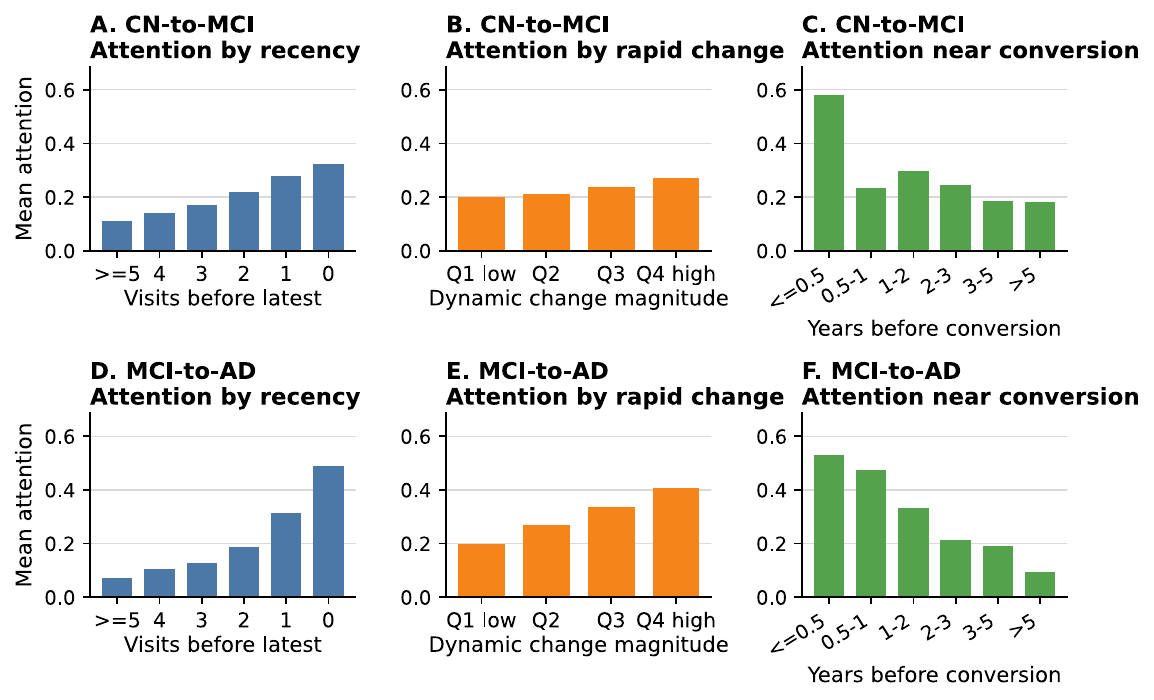}%
    }{%
        \missingfigure{figure/figure5.pdf}%
    }
    \caption{Visit-level attention context by recency, dynamic change magnitude, and conversion proximity.}
    \label{fig:figure5}
\end{figure}

Attention also emphasized recent and conversion-proximal visits (Figure~\ref{fig:figure5}). Latest-visit attention exceeded earlier-visit attention in both tasks, and attention correlated with index proximity, conversion proximity among converters, and dynamic change magnitude, with stronger associations for MCI-to-AD.

\begin{table}[t]
\centering
\scriptsize
\setlength{\tabcolsep}{2pt}
\caption{MCI-to-AD ablation study on the held-out test set. IBS is lower-is-better; all other metrics are higher-is-better.}
\label{tab:ablation-study}
\resizebox{\columnwidth}{!}{%
\begin{tabular}{lccccc}
\hline
Variant & C-index & IBS $\downarrow$ & Mean TD AUC & 5y AUROC & 5y AUPRC \\
\hline
Full PROMISE-AD & \best{$0.894\pm0.011$} & \best{$0.095\pm0.007$} & $0.922\pm0.012$ & \best{$0.997\pm0.003$} & \best{$0.999\pm0.001$} \\
No dynamic features & $0.870\pm0.017$ & $0.103\pm0.005$ & \best{$0.926\pm0.004$} & $0.984\pm0.011$ & $0.992\pm0.003$ \\
No visit dropout & $0.879\pm0.020$ & $0.104\pm0.008$ & $0.908\pm0.018$ & $0.990\pm0.005$ & $0.994\pm0.001$ \\
No representation fusion & $0.879\pm0.002$ & $0.106\pm0.001$ & $0.906\pm0.011$ & $0.985\pm0.009$ & $0.993\pm0.002$ \\
No mixture hazards & $0.880\pm0.018$ & $0.107\pm0.003$ & $0.903\pm0.021$ & $0.986\pm0.002$ & $0.993\pm0.001$ \\
No progression ranking & $0.882\pm0.013$ & $0.104\pm0.007$ & $0.913\pm0.012$ & $0.991\pm0.005$ & $0.994\pm0.001$ \\
No horizon loss & $0.882\pm0.011$ & $0.104\pm0.007$ & $0.911\pm0.013$ & $0.991\pm0.003$ & $0.994\pm0.001$ \\
\hline
\end{tabular}
}
\end{table}

\section{Discussion}
PROMISE-AD provides a leakage-safe longitudinal survival framework for estimating multi-horizon AD conversion risk from irregular pre-index visits. Its novelty lies in an AD-specific integration rather than a new Transformer or survival loss alone: the framework joins leakage-safe cohort construction, progression-aware incomplete-record tokenization, hybrid temporal fusion, latent mixture hazards, and calibrated horizon risks in one supervised conversion-time model. The two tasks exposed different prediction regimes: CN-to-MCI was sparse and late-converting, with only one test conversion before 1 year and six before 5 years. This is a fundamental dataset constraint that makes CN-to-MCI fixed-horizon discrimination and calibration exploratory rather than definitive. MCI-to-AD had more frequent events and yielded stronger risk stratification. Across these settings, PROMISE-AD linked temporal representation learning with survival objectives, censoring-aware risk estimation, and horizon calibration.

The comparison with baselines shows that PROMISE-AD should be interpreted as a balanced risk-estimation framework rather than a uniformly dominant classifier. It produced the lowest CN-to-MCI IBS, supporting improved time-integrated probability estimation in the lower-event task. For MCI-to-AD, it achieved the best C-index and near-ceiling 5-year discrimination, while CoxTime achieved the lowest IBS. This distinction matters clinically because ranking, calibration, and horizon-specific discrimination support different decisions.

Ablation and interpretability analyses supported the technical design. The full model provided the best C-index, IBS, and 5-year ablation metrics, while dynamic change/slope features, representation fusion, and latent mixture hazards contributed most clearly to ranking, probability error, and late-horizon discrimination. The mixture head is clinically relevant as a way to represent heterogeneous progression speeds, because its experts define soft rapid, delayed, and lower-risk hazard patterns rather than forcing all subjects to share one hazard shape. Attributions emphasized memory, cognitive, functional, temporal, entorhinal, parahippocampal, and hippocampal measures, and attention favored recent, conversion-proximal, and dynamically changing visits, aligning the learned representation with plausible disease progression signals.

Several limitations remain. The study used ADNI/TADPOLE, a well-characterized research cohort that may not represent community memory-clinic or primary-care populations; external validation remains essential \cite{steyerberg2019clinical,collins2015tripod}. A central limitation is the sparse CN-to-MCI test event count, especially the single conversion before 1 year and six conversions before 5 years, which makes horizon-specific AUROC, AUPRC, and calibration estimates sensitive to individual subjects. These CN-to-MCI horizon metrics should therefore be viewed as preliminary evidence constrained by the dataset, not as stable early-conversion performance estimates. Late MCI-to-AD nonconversion was also limited. In addition, the mixture experts were not externally validated against biomarker-defined or neuropathological AD subtypes, so they should not be interpreted as confirmed clinical phenotypes. Future work should test additional cohorts, temporal validation schemes, broader biomarker and clinical variables, expert-level subgroup profiling, comparison with established AD subtype/staging models, and site-specific calibration. Finally, the interpretability analyses identify associations with predicted risk rather than causal effects.


\bibliographystyle{IEEEtran}
\bibliography{references}

\end{document}